\documentclass{svmult}

\usepackage{mathptmx}       
\usepackage{helvet}         
\usepackage{courier}        
\usepackage{type1cm}        
\usepackage{makeidx}         
\usepackage{graphicx}        
\usepackage{multicol}        
\usepackage[bottom]{footmisc}
\usepackage{amsmath,amssymb,epsfig}
\usepackage{psfrag}
\usepackage{amsmath}
\usepackage{array,contour}
\newcommand{\gbf}[1] {\contour[5]{black}{${#1}$}}

\makeindex             

\begin{document}
\title*{Workspace and Singularity analysis of a Delta like family robot}
\author{R. Jha$^1$, D. Chablat$^1$,  F. Rouillier$^2$ and G. Moroz$^3$}
\authorrunning{R. Jha et al.}
\institute{$^1$Institut de Recherche en Communications et Cybern\'etique de Nantes,\\ UMR CNRS 6597, France,\email{Ranjan.Jha,Damien.Chablat@irccyn.ec-nantes.fr} \\
$^2$INRIA Paris-Rocquencourt, Institut de Math\'ematiques de Jussieu,\\ UMR CNRS 7586, France,\email{Fabrice.Rouillier@inria.fr} \\
$^3$INRIA Nancy-Grand Est, France,\email{Guillaume.Moroz@inria.fr}}
\maketitle
\abstract{Workspace and joint space analysis are essential steps in describing the task and designing the control loop of the robot, respectively. This paper presents the descriptive analysis of a family of delta-like parallel robots by using algebraic tools to induce an estimation about the complexity in representing the singularities in the workspace and the joint space. A Gr\"{o}bner based elimination is used to compute the singularities of the manipulator and  a Cylindrical Algebraic Decomposition algorithm is used to study the workspace and the joint space. From these algebraic objects, we propose some certified three dimensional plotting describing the the shape of workspace and of the joint space which will help the engineers or researchers to decide the most suited configuration of the manipulator they should use for a given task. Also, the different parameters associated with the complexity of the serial and parallel singularities are tabulated, which further enhance the selection of the different configuration of the manipulator by comparing the complexity of the singularity equations.}
\section{Introduction}
The workspace can be defined as the volume of space or the complete set of poses which the end-effector of the manipulator can reach. Many researchers published several works on the problem of computing these complete sets for robot kinematics. Based on the early studies \cite{soni:1983, Hansen:1983}, several methods for workspace determination have been proposed, but many of them are applicable only for a particular class of robots. The workspace of parallel robots mainly depends on the actuated joint variables, the range of motion of the joints and the mechanical interferences between the bodies of mechanism. There are different techniques based on geometric \cite{Gosselin:1990, Merlet:1992}, discretization \cite{castelli:2008, Ilian:2001, Chablat:2004}, and algebraic methods \cite{zein:2006,ottaviano:2006, Chablat:2014, Chablat:2011} which can be used to compute the workspace of parallel robot. The main advantage of the geometric approach is that, it establish the nature of the boundary of the workspace \cite{Siciliano:2008}. Also it allows to compute the surface and volume of the workspace while being very
efficient in terms of storage space, but when the rotational motion is included, it becomes less efficient.\\
Interval analysis based methods can be used to compute the workspace but the computation time depends on the complexity of the robot and the requested accuracy \cite{Chablat:2004}.  Discretization methods are usually less complicated and can easily take into account all kinematic constraints, but they require more space and computation time for higher resolutions. The majority of numerical methods which is used to determine the workspace of parallel manipulators includes the discretization of the pose parameters for computing workspace boundaries \cite{Ilian:2001}. There are other approaches, which are based on optimization algorithms \cite{snyman:2000} for fully serial or parallel manipulators, analytic methods for symmetrical spherical mechanisms \cite{bonev:2006}. In \cite{bohigas:2012} a method for computing the  workspace boundary for manipulators with a general structure is proposed, which uses a branch-and-prune technique to isolate a set of output singularities, and then classifies the points on such set according to whether they correspond to motion impediments in the workspace. A cylindrical algebraic decomposition (CAD) based method is illustrated in \cite{Chablat:2014,chab:2014}, which is used to model the workspace and joint space for the 3 RPS parallel robot.

This paper presents the results which are obtained by applying algebraic methods for the workspace and joint space analysis of a family of delta-like robot including complexity information for representing the singularities in the workspace and the joint space. The CAD algorithm is used to study the workspace and joint space, and a Gr\"{o}bner based elimination process is used to compute the parallel and serial singularities of the manipulator. The structure of the paper is as follows. Section 2 describes the architecture of the manipulator, including kinematic equation and joint constraints associated with the manipulators. Section 3 discusses the computation of parallel as well as serial singularities and their projections in workspace and joint-space. Section 4 presents a comparative study on the shape of the workspace of different delta-like robots. Section 5 finally concludes the paper.
\section{Manipulators Architecture}
The manipulator architecture under the study is a three degree of freedom parallel mechanism which consists of three identical legs, the different arrangements of these legs give rise to family of delta like robot. Several types of delta-like robot were studied, few of them are Orthoglide \cite{Chablat:2004,Pashkevich:2006}, Hybridglide, Triaglide \cite{hebsacker:1999} and UraneSX \cite{Chablat:2004}. The position vectors of end points of $i^{th}$ leg are  ${\bf P}^{j}$ and ${\bf B}_i^{j}$, also ${\bf A}_i^{j}$ and ${\bf B}_i^{j}$ are the position vectors of end points of $i^{th}$ actuator where $j$ represents the manipulator type from the family of delta like robot ($j$ = 1 - Orthoglide, 2 - Hybridglide, 3 - Triaglide, 4 - UraneSX). ${\gbf \rho}_i^{j}$ represents the prismatic joint variables whereas ${\bf P}^{j}$ represents the position vector of the tool center point which is shown in Eq. (\ref{eq:actuator}).
\begin{equation}\label{eq:actuator}
||{\bf A}_i^{j}{\bf B}_i^{j}||= \rho_i^{j} \quad  {\bf P}= [x~y~z]^T \quad {\rm with} \quad i=1, 2, 3 \quad {\rm and} \quad j=1, 2, 3, 4
\end{equation}

\begin{figure}[!ht]
    \begin{center}
    \begin{tabular}{@{}c@{}c@{}}
       \begin{minipage}[t]{45 mm}%
		    \psfrag{p1}{\small $P_1$}
        \psfrag{p2}{\small $P_2$}
				\psfrag{a1}{\small $A_1$}
        \psfrag{a2}{\small $A_2$}
				\psfrag{a3}{\small $A_3$}
				\psfrag{b1}{\small $B_1$}
        \psfrag{b2}{\small $B_2$}
				\psfrag{b3}{\small $B_3$}
       	\includegraphics[width=32 mm]{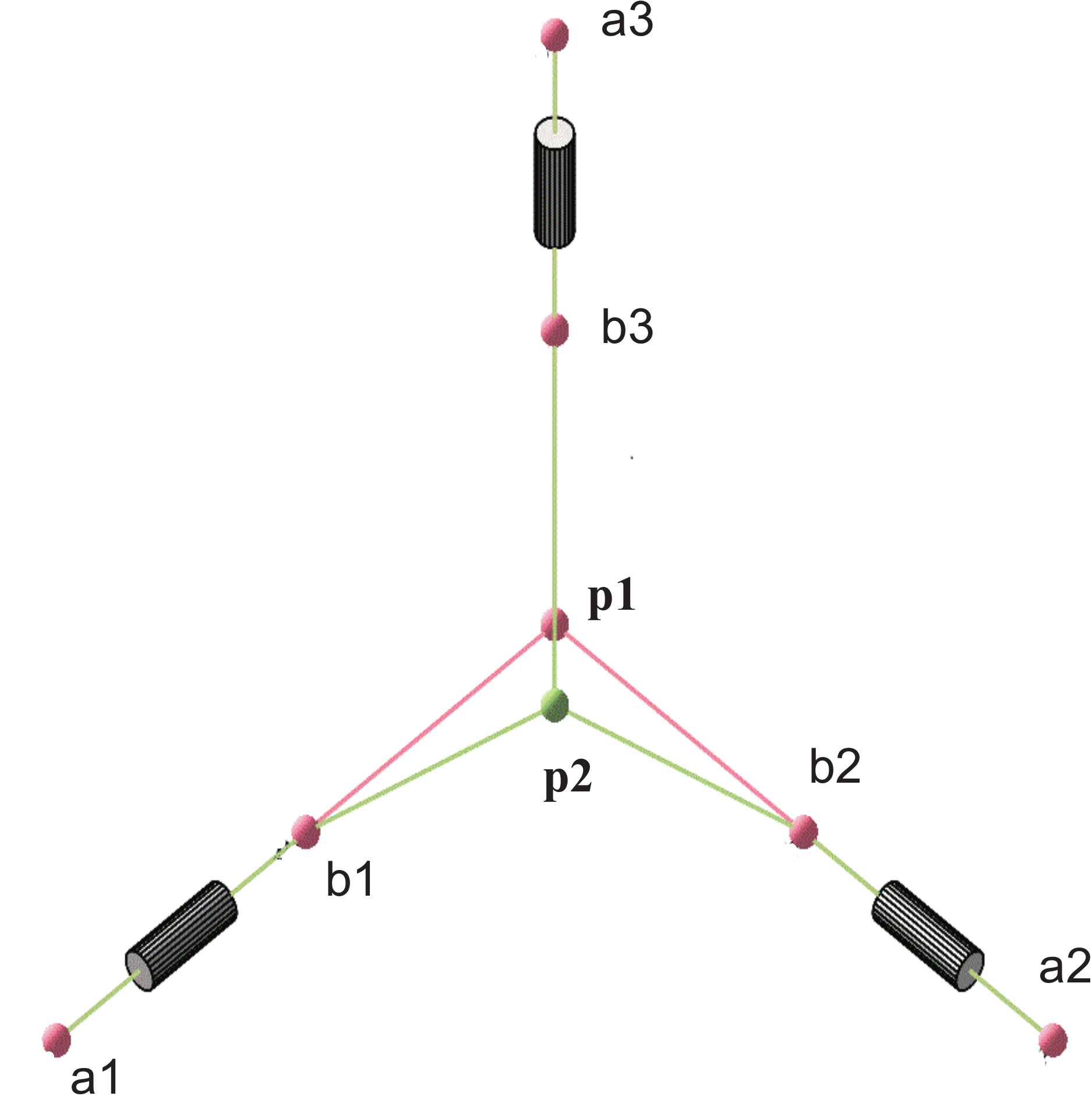}(a)
       \end{minipage} &
       \begin{minipage}[t]{45 mm}%
		    \psfrag{p1}{\small $P_1$}
        \psfrag{p2}{\small $P_2$}
				\psfrag{a1}{\small $A_1$}
        \psfrag{a2}{\small $A_2$}
				\psfrag{a3}{\small $A_3$}
				\psfrag{b1}{\small $B_1$}
        \psfrag{b2}{\small $B_2$}
				\psfrag{b3}{\small $B_3$}
        \includegraphics[width=32 mm]{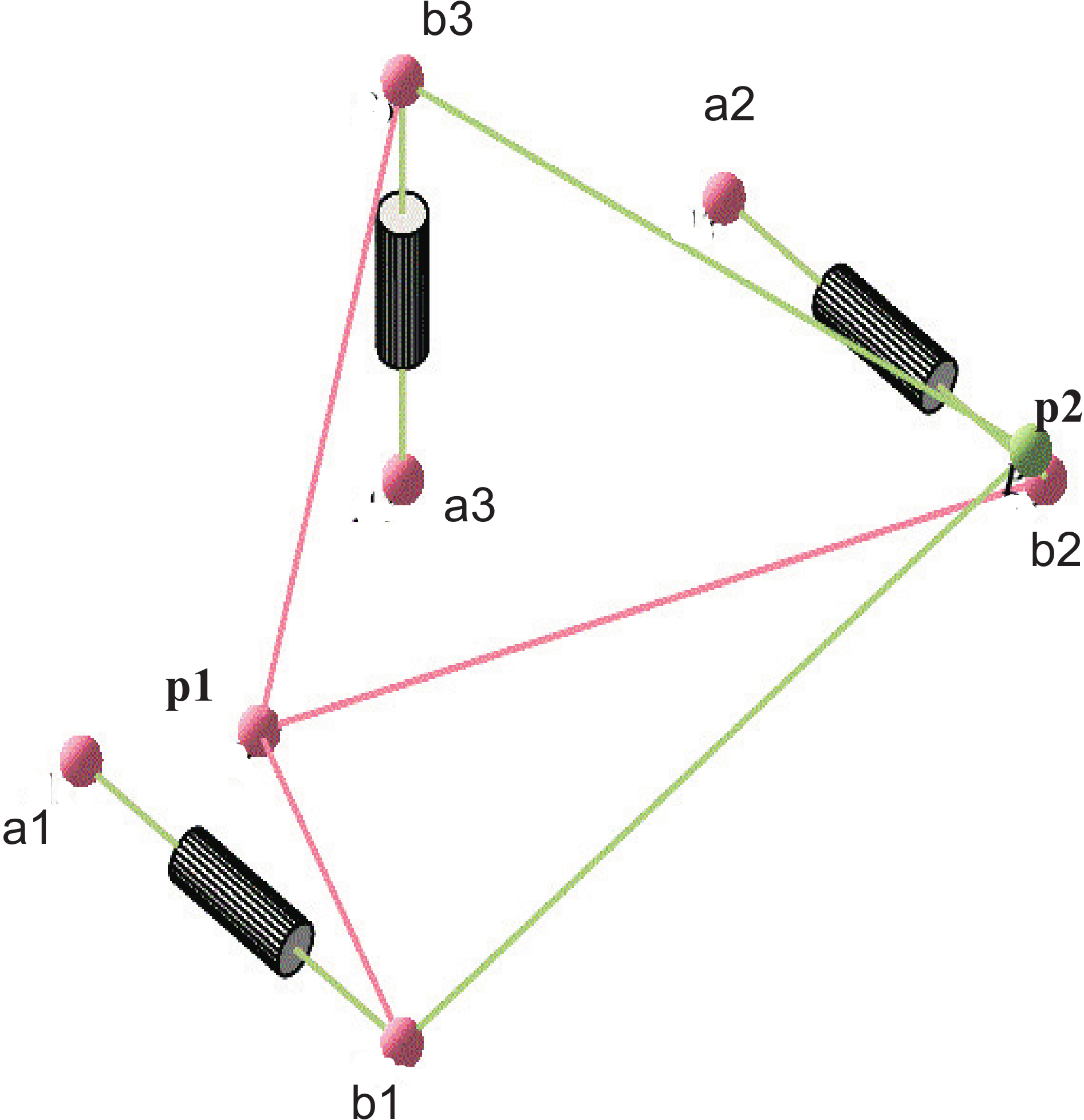}(b)
       \end{minipage} \\
			 \begin{minipage}[t]{45 mm}%
		    \psfrag{p1}{\small $P_1$}
        \psfrag{p2}{\small $P_2$}
				\psfrag{a1}{\small $A_1$}
        \psfrag{a2}{\small $A_2$}
				\psfrag{a3}{\small $A_3$}
				\psfrag{b1}{\small $B_1$}
        \psfrag{b2}{\small $B_2$}
				\psfrag{b3}{\small $B_3$}
        \includegraphics[width=32 mm]{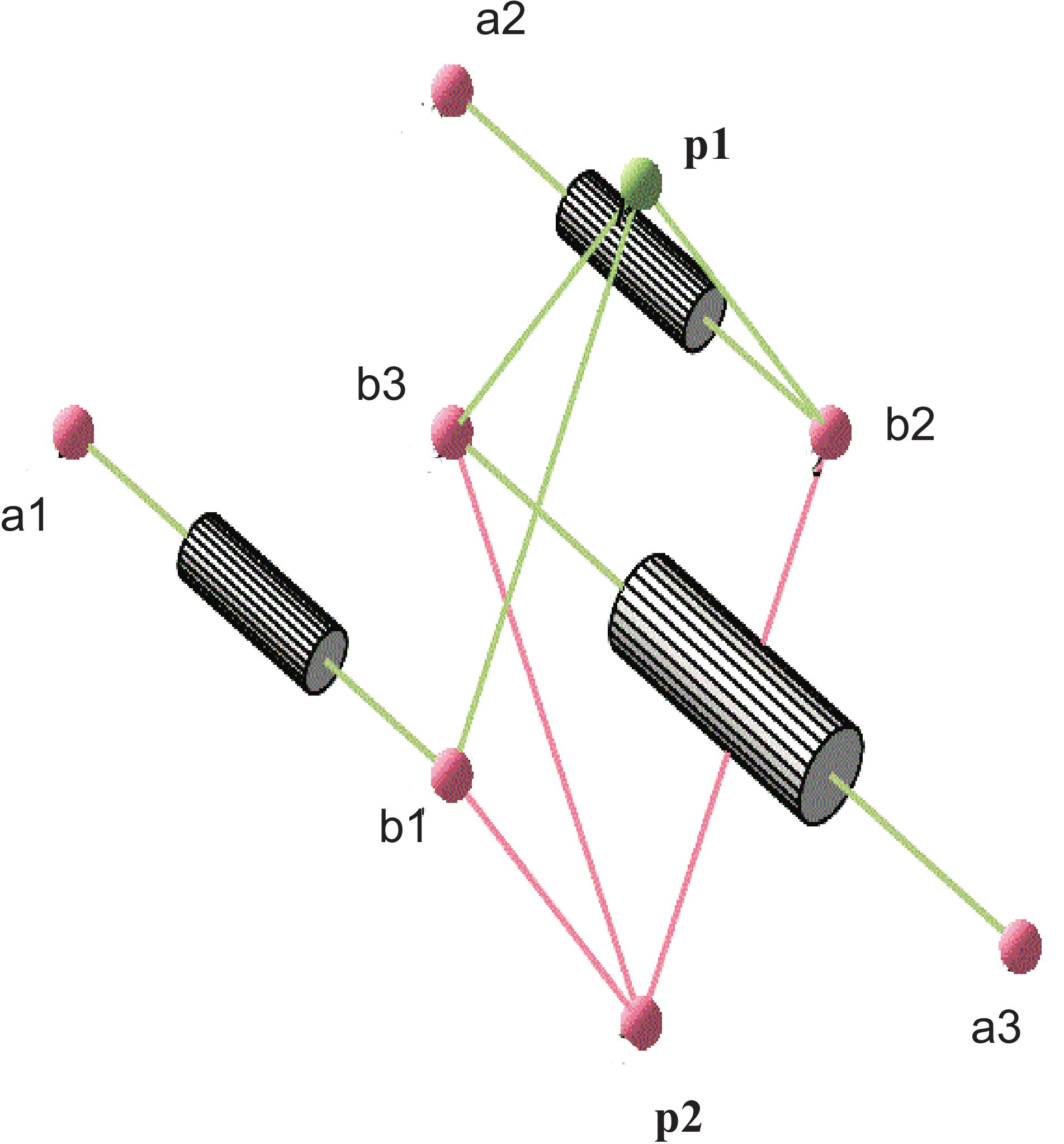}(c)
       \end{minipage} &
       \begin{minipage}[t]{45 mm}%
		    \psfrag{p1}{\small $P_1$}
        \psfrag{p2}{\small $P_2$}
				\psfrag{a1}{\small $A_1$}
        \psfrag{a2}{\small $A_2$}
				\psfrag{a3}{\small $A_3$}
				\psfrag{b1}{\small $B_1$}
        \psfrag{b2}{\small $B_2$}
				\psfrag{b3}{\small $B_3$}
        \includegraphics[width=32 mm]{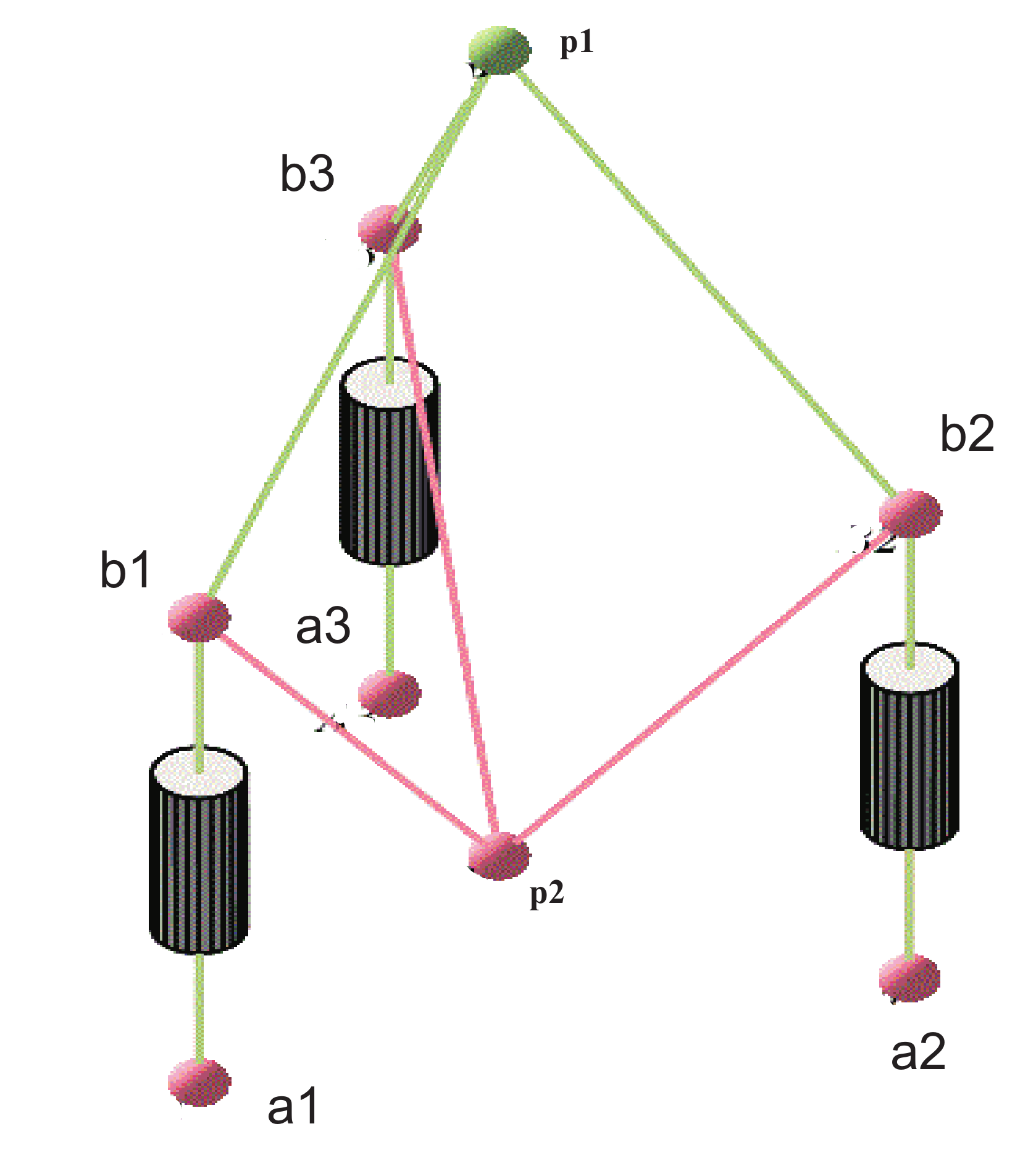}(d)
       \end{minipage}
			    \end{tabular}
    \end{center}
    \caption[Configuration plot for delta robots]{Configuration plot for Orthoglide (a), Hybridglide (b), Triaglide (c) and UraneSX (d) robot}
    \protect\label{figure:configuration}
\end{figure}
The kinematic equations of the family of delta like robot can be generalized as
$||{\bf P} - {\bf B}^{j}_i||= {{\bf L}^{j}_i}$.  
%
All the computations and analysis are done for ${{\bf L}^{j}_i} = {\bf L}$ and by imposing the following constraints on joint variables. Without joint limits, all the robots admit two assembly modes and eight working modes.
\begin{equation}\label{eq:joint_constraints}
0 < \rho_1 < 2L \quad 0 < \rho_2 < 2L \quad 0 < \rho_3 < 2L
\end{equation}
The {\it Orthoglide} mechanism is driven by three actuated orthogonal prismatic joints. A simpler virtual model can be defined for the Orthoglide, which consists of three bar links connected by the revolute joints to the tool center point on one side and to the corresponding prismatic joint at another side. Several assembly modes of these robots depends upon the solutions of direct kinematic problem is shown in Fig.~\ref{figure:configuration}(a). The point ${\bf P}_i$ represents the pose of corresponding robot. However more than one value of $i$ for the point ${\bf P}_i$ shows the multiple solutions for the DKP.
The constraint equations for the Orthoglide are
\begin{equation}\label{eq:orthoglide}
(x-\rho_1)^2\! + \! y^2 \!+ \! z^2 \!\! = \!\! L^2;
 x^2 \!+\! (y-\rho_2)^2 \!+ \!z^2 \!\! = \!\! L^2;
 x^2 \!+ \!y^2 \!+ \!(z-\rho_3)^2 \!\! = \!\! L^2 \nonumber
\end{equation}
The {\it Hybridglide} mechanism consists of three actuated prismatic joints, in which two actuators are placed parallel and third one perpendicular to others two.  Also the three bar links connected by spherical joints to the tool center point on one side and to the corresponding prismatic joint at another side. Several assembly modes of these robots depends upon the solutions of direct kinematic problem is shown in Fig.~\ref{figure:configuration}(b) and following is the constraint equations:
\begin{equation}\label{eq:hybridglide}
(x-1)^2 + (y-\rho_1)^2 + z^2  = L^2; 
(x+1)^2 + (y-\rho_2)^2 + z^2  = L^2;  
x^2 + y^2 + (z-\rho_3)^2  = L^2 \nonumber
\end{equation}
The {\it Triaglide} manipulator is driven by three actuated prismatic joints, in which all the three actuators parallel to each other and placed in the same plane. The architecture of the Triaglide is shown in Fig.~\ref{figure:configuration}(c) and the constraint equations are defined as:
\begin{equation}\label{eq:triaglide}
(x-1)^2 + (y-\rho_1)^2 + z^2 \!= \!L^2;
(x+1)^2 + (y-\rho_2)^2 + z^2 \!= \!L^2;
x^2 + (y-\rho_3)^2 + z^2 \!= \!L^2\nonumber
\end{equation}
The {\it UraneSX} is similar to triaglide, but instead of three actuators in the same plane, it is placed in different planes. The architecture of the Triaglide is shown in Fig.~\ref{figure:configuration}(d) and the constraint equations are as follows:
\begin{eqnarray}\label{eq:uranesx}
&&(x-1)^2 + y^2+(z-\rho_1)^2 = L^2;
(x+1/2)^2 + (y-\sqrt{3}/2)^2 + (z-\rho_2)^2 \!= \! L^2;\nonumber \\
&&(x+1/2)^2 + (y+\sqrt{3}/2)^2 + (z-\rho_3)^2 \! = \! L^2\nonumber
\end{eqnarray}
\section{Singularity Analysis}
Singularities of a robotic manipulator are important feature that essentially influence its capabilities. Mathematically, a singular configuration may be defined a rank deficiency of the Jacobian describing the differential mapping from the joint space to the workspace and vice versa. Differentiating the constraint equations of the robot with respect to time leads to the velocity model:
 $ {\bf A} {\bf t}+ {\bf B} \dot{\bf q}= \bf{0}$
where $\bf A$ and $\bf B$ are the parallel and serial Jacobian matrices, respectively, ${\bf t}$ is the velocity of ${\bf P}$ and $\dot{\bf q}$ defines the joint velocities. The parallel singularities occur whenever ${\rm det}({\bf A})=0$ and the serial singularities occur whenever ${\rm det}({\bf B})=0$ \cite{chablat:1998}.
\begin{align}
\label{eq:singlr}
{{\rm det}({\bf A})_o} &= -8\rho_1\rho_2\rho_3 + 8\rho_1\rho_2{z} + 8\rho_1\rho_3{y} + 8\rho_2\rho_3{x} \nonumber \\
{{\rm det}({\bf A})_h} &= -8\rho_1\rho_3{x}+8\rho_2\rho_3{x}-8\rho_1\rho_3+8\rho_1{z}-8\rho_2\rho_3+8\rho_2{z}+16\rho_3{y} \nonumber \\
{{\rm det}({\bf A})_t} &=  8\rho_1{z}+8\rho_2{z}-16\rho_3{z} \nonumber \\
{{\rm det}({\bf A})_u} &=  4\sqrt{3}(3z-\rho_1-\rho_2-\rho_3+\rho_3x+\rho_2x-2\rho_1x)+12\rho_3y-12\rho_2y
\end{align}
Parallel and serial singularities as well as their projections in workspace and joint space are computed using a Gr\"{o}bner based elimination method. This usual way for eliminating variables (see \cite{LCO04}) computes (the algebraic closure of) the projection of the parallel singularities in the workspace. In the same way, one can compute (the algebraic closure of) the projection of the parallel singularities in the joint space. Both are then defined as the zero set of some system of algebraic equations and we assume that the considered robots are generic enough so that both are hypersurfaces. ${{\rm det}({\bf A})_o}$, ${{\rm det}({\bf A})_h}$, ${{\rm det}({\bf A})_t}$ and ${{\rm det}({\bf A})_u}$ are the parallel singularities of Orthoglide, Hybridglide, Triaglide and UraneSX, respectively. Starting from the constraint equations and the determinant of the Jacobian matrix, we are able to eliminate the joint values. This elimination strategy is more efficient than a cascading elimination by means of resultants which might introduce many more spurious solutions : singular points that are not projections of singular points. Due to the lack of space, other equations associated with serial singularities are not presented in this paper.
\begin{figure}[!hb]
    \begin{center}
    \begin{tabular}{@{}c@{}c@{}}
       \begin{minipage}[t]{45mm}%
		    \psfrag{x}{$x$}
				\psfrag{y}{$y$}
				\psfrag{z}{$z$}
        \includegraphics[width=32mm]{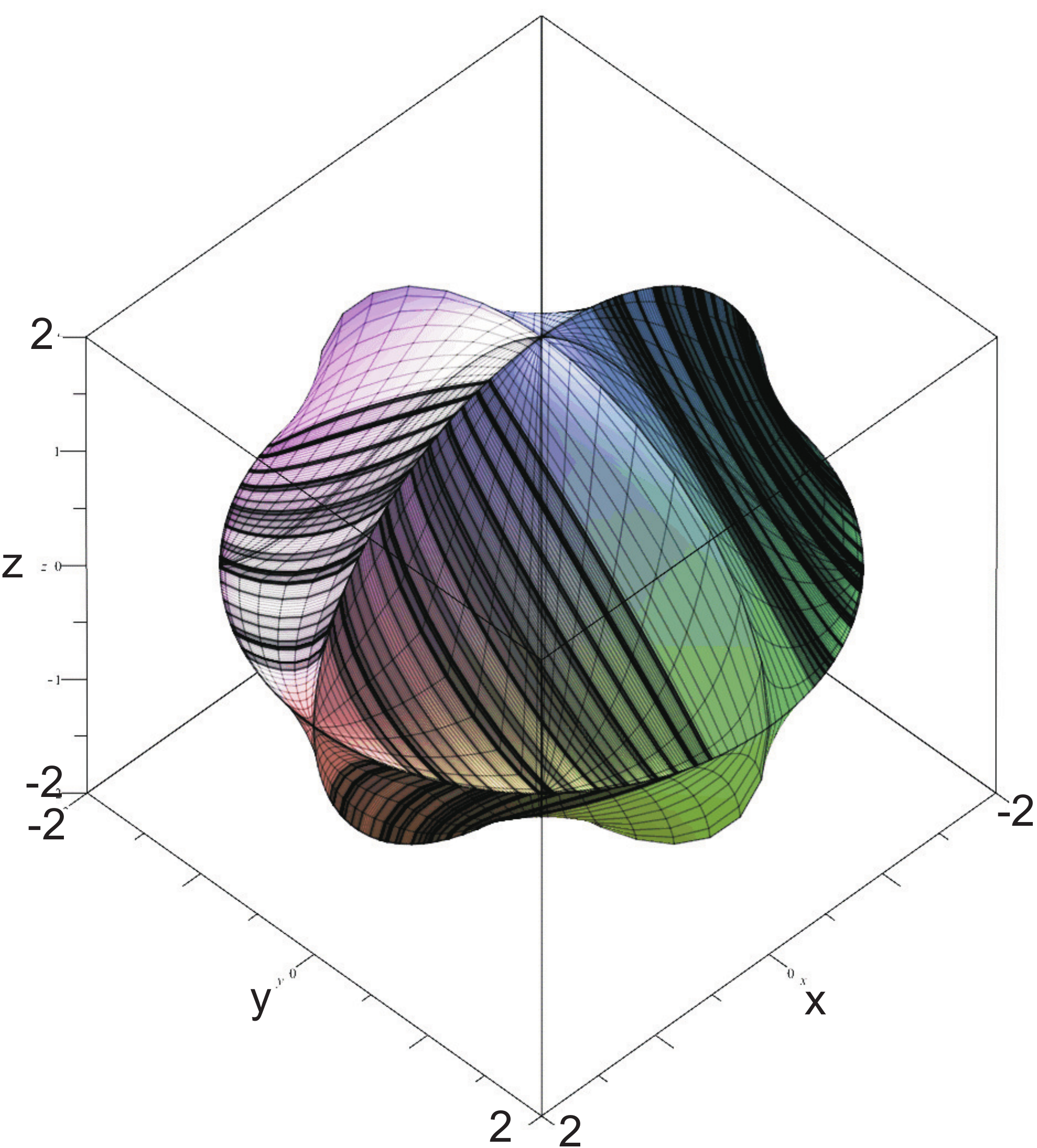}(a)
       \end{minipage} &
       \begin{minipage}[t]{45mm}%
		    \psfrag{x}{$x$}
				\psfrag{y}{$y$}
				\psfrag{z}{$z$}
        \includegraphics[width=32mm]{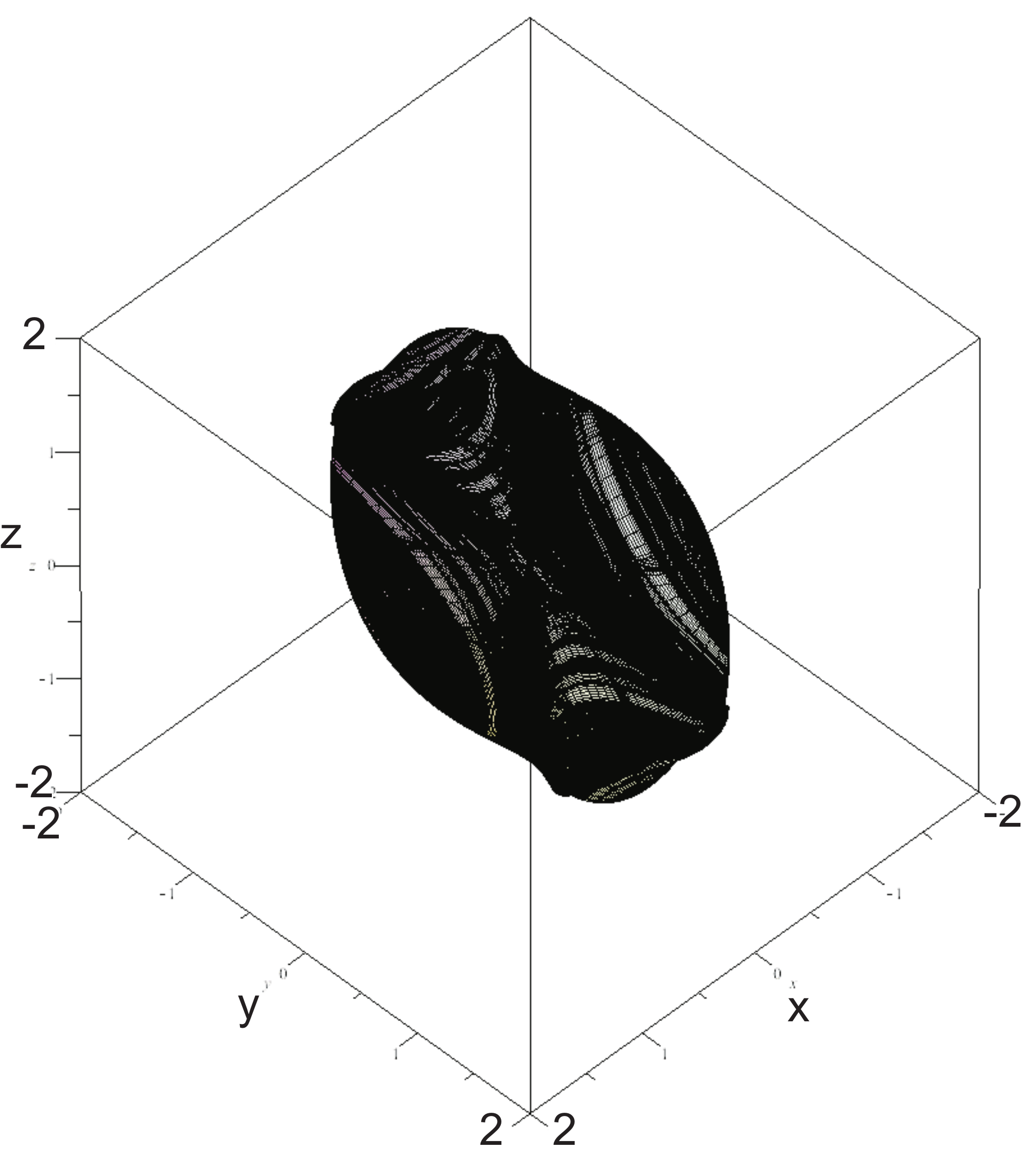}(b)
       \end{minipage} \\
       \begin{minipage}[t]{45mm}%
		    \psfrag{x}{$x$}
				\psfrag{y}{$y$}
				\psfrag{z}{$z$}
        \includegraphics[width=32mm]{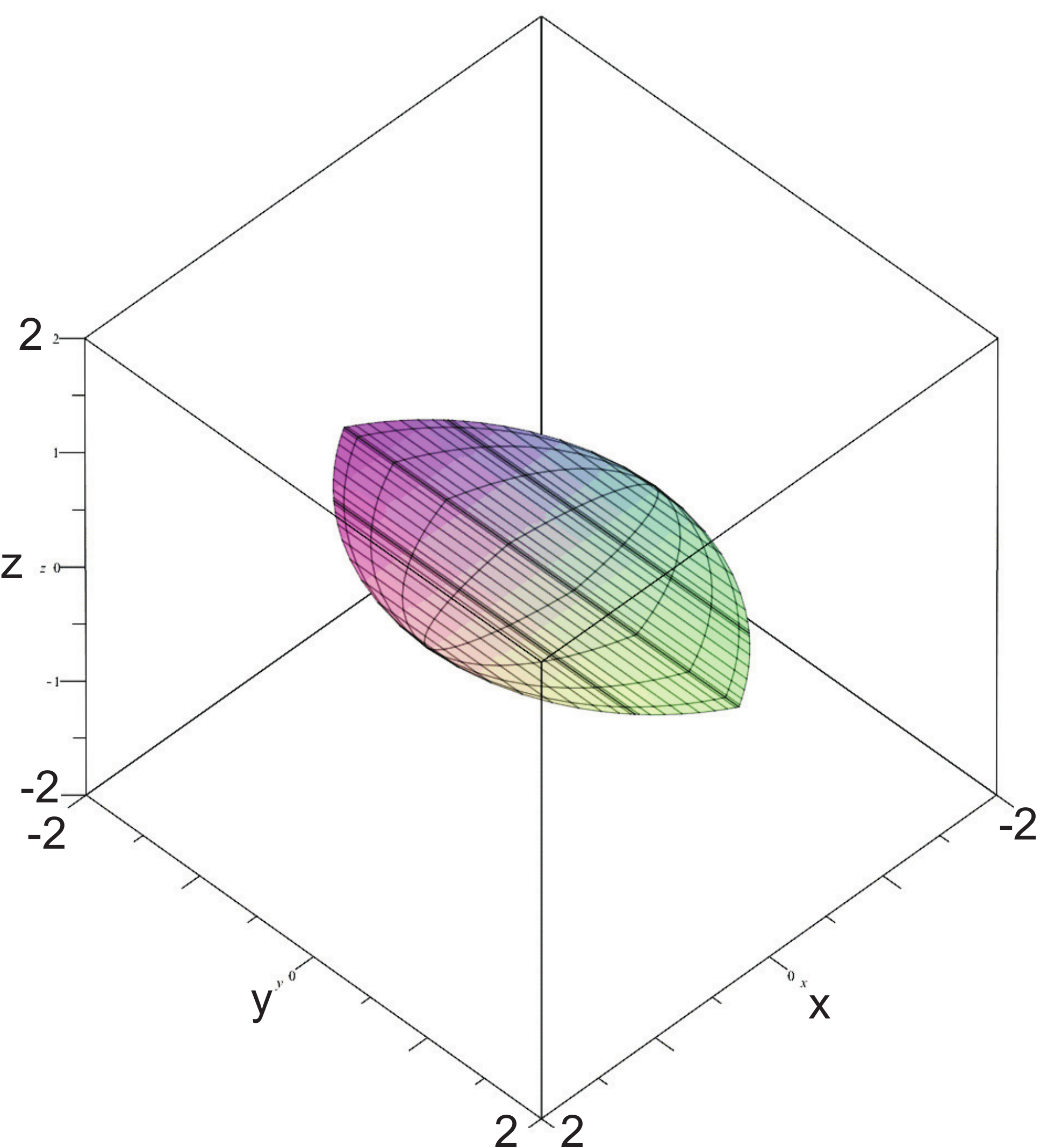}(c)
       \end{minipage} &
       \begin{minipage}[t]{45mm}%
		    \psfrag{x}{$x$}
				\psfrag{y}{$y$}
				\psfrag{z}{$z$}
        \includegraphics[width=32mm]{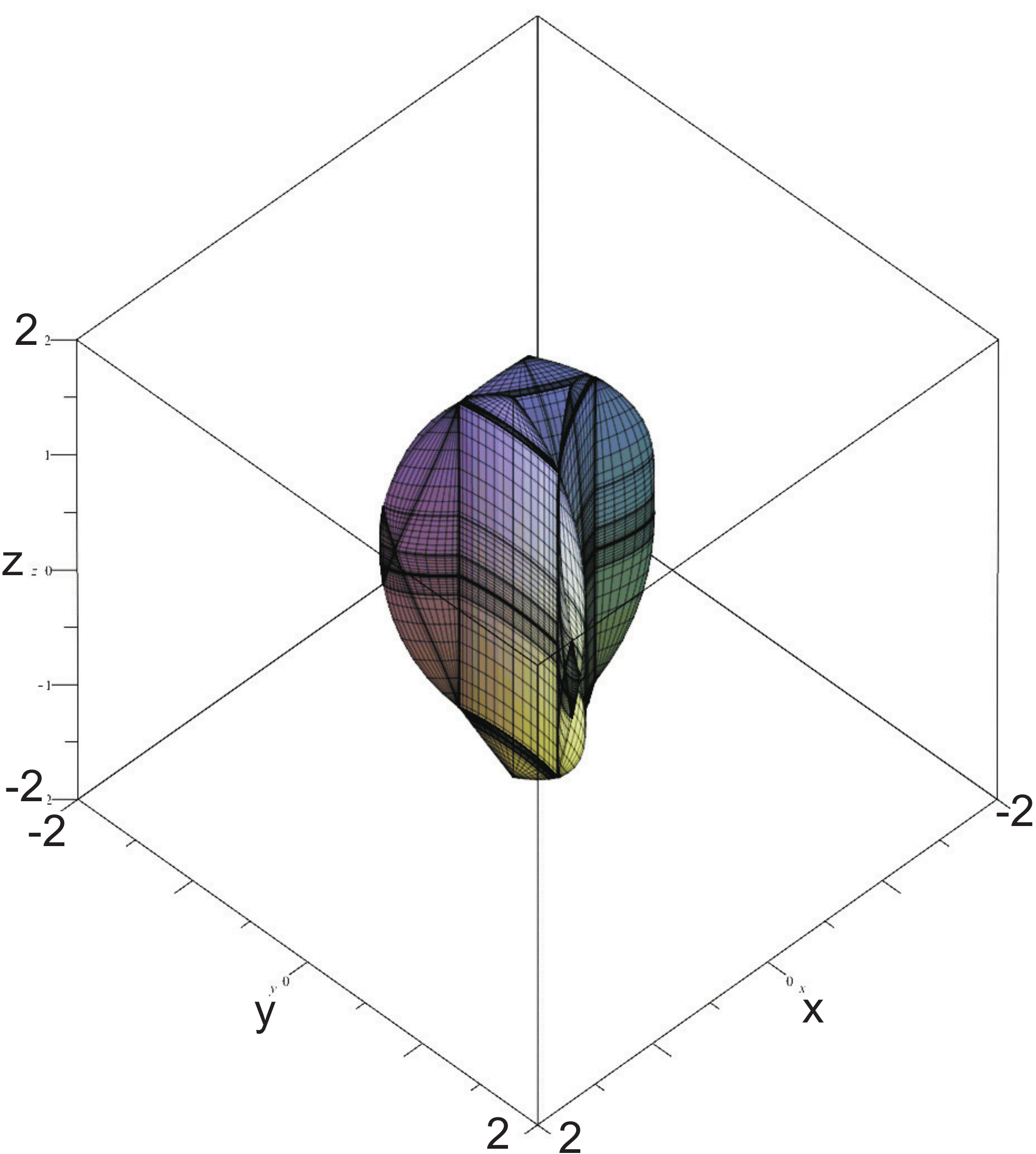}(d)
       \end{minipage}
    \end{tabular}
    \end{center}
    \caption[Projection of parallel singularities in workspace]{Projection of parallel singularities of Orthoglide (a), Hybridglide (b), Triaglide (c) and UraneSX (d) in workspace. All the computations are done with joint limits.}
\protect\label{figure:parallel_ws}
\end{figure}
Figure~\ref{figure:parallel_ws} represents the projections of parallel singularities in projection space $({\bf x,  y, z})$. These surfaces represent the singularity associated with the eight working modes.
\begin{table}
\caption{Comparison of the different parameters associated with the singularities for the robots}
\label{complex}
\centering
\begin{tabular}{l l c c c c r}
\svhline\noalign{\smallskip}
Manipulators & Types of      & Plotting Time & Degrees & No.of terms & Binary  & No. of Cells  \\
             & Singularities & ($s$)  &         &             & Size    &                \\
\hline\noalign{\smallskip}
Orthoglide  & Parallel & 0037.827  & 18[10,10,10] & 097 & 015 & [02382,0272] \\
            & Serial   & 0005.133  & 18[12,12,12] & 062 & 012 & [00044,0004]\\
Hybridglide & Parallel & 5698.601  & 20[16,08,12] & 119 & 017 & [28012,1208]\\
            & Serial   & 0007.007  & 18[12,12,12] & 281 & 017 & [00158,0027]\\
Triaglide   & Parallel & 0010.625  & 03[00,00,03] & 002 & 002 & [00138,0004]\\
            & Serial   & 0005.079  & 06[06,06,06] & 042 & 007 & [00077,0017]\\
UraneSX     & Parallel & 0022.625  & 06[06,04,00] & 015 & 040 & [02795,0070]\\
            & Serial   & 0018.391  & 12[12,12,12] & 252 & 151 & [00392,0142]\\
\svhline\noalign{\smallskip}
\end{tabular}
\end{table}

In Table~\ref{complex},  a comparative study of five parameters among the family of delta like robot is presented. We have tabulated the main characteristics of the polynomials (In three variables) used for the plots (Implicit surface) : their total degree, their number of terms and the maximum bitsize of their coefficients. We have also reported the time (In seconds) for plotting the implicit surface which they define and the number of cells computed by the CAD, as well as the number of cells in the final result after gluing those that are adjacent and belongs to the same connected component. Several functions are used which involves the discriminant variety, Gr\"{o}bner bases and CAD computations, computed in Maple 18 with a Intel(R) Core(TM) i7-3770 CPU @ 3.40 GHz (14 Gb RAM).  As can be seen from Table~\ref{complex},  there exists higher values of all the parameters for the Hybridglide, among all manipulators listed, which infers that it has more complex singularities, whereas for the Triaglide all the values are least which intuits the less complicated singularities. For example, the computation times for the Hybridglyde for parallel singularities is high compared to the one for the Othoglide, even if the surface has similar characteristics. This is due to the geometry of the surface which is more difficult to decompose in the case of the Hybridglide : the Cylindrical Algebraic Decomposition is described  by 1208 cylindrical cells in the case of Hybridglide while it is described by 272 cells for the Orthoglide.

\section{Workspace analysis}
The workspace analysis allows to characterize of the workspace regions where the number of real solutions for the inverse kinematics is constant. A CAD algorithm is used to compute the workspace of the robot in the projection space $(x, y, z)$ with some joint constraints taken in account.
\begin{figure}[!ht]
    \begin{center}
			   \psfrag{1 Solns}{1 Solns}
				 \psfrag{2 Solns}{2 Solns}
				 \psfrag{4 Solns}{4 Solns}
				 \psfrag{8 Solns}{8 Solns}
  			\includegraphics[scale = 0.25]{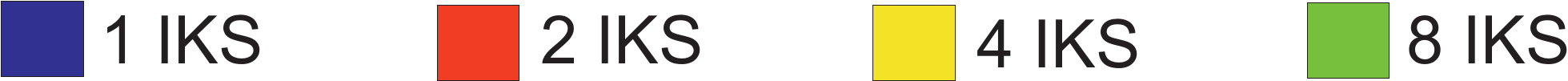} \\
        \begin{tabular}{@{}c@{}c@{}}
		     \begin{minipage}[t]{45 mm}%
		    \psfrag{x}{$x$}
				\psfrag{y}{$y$}
				\psfrag{z}{$z$}
        \includegraphics[width=32 mm]{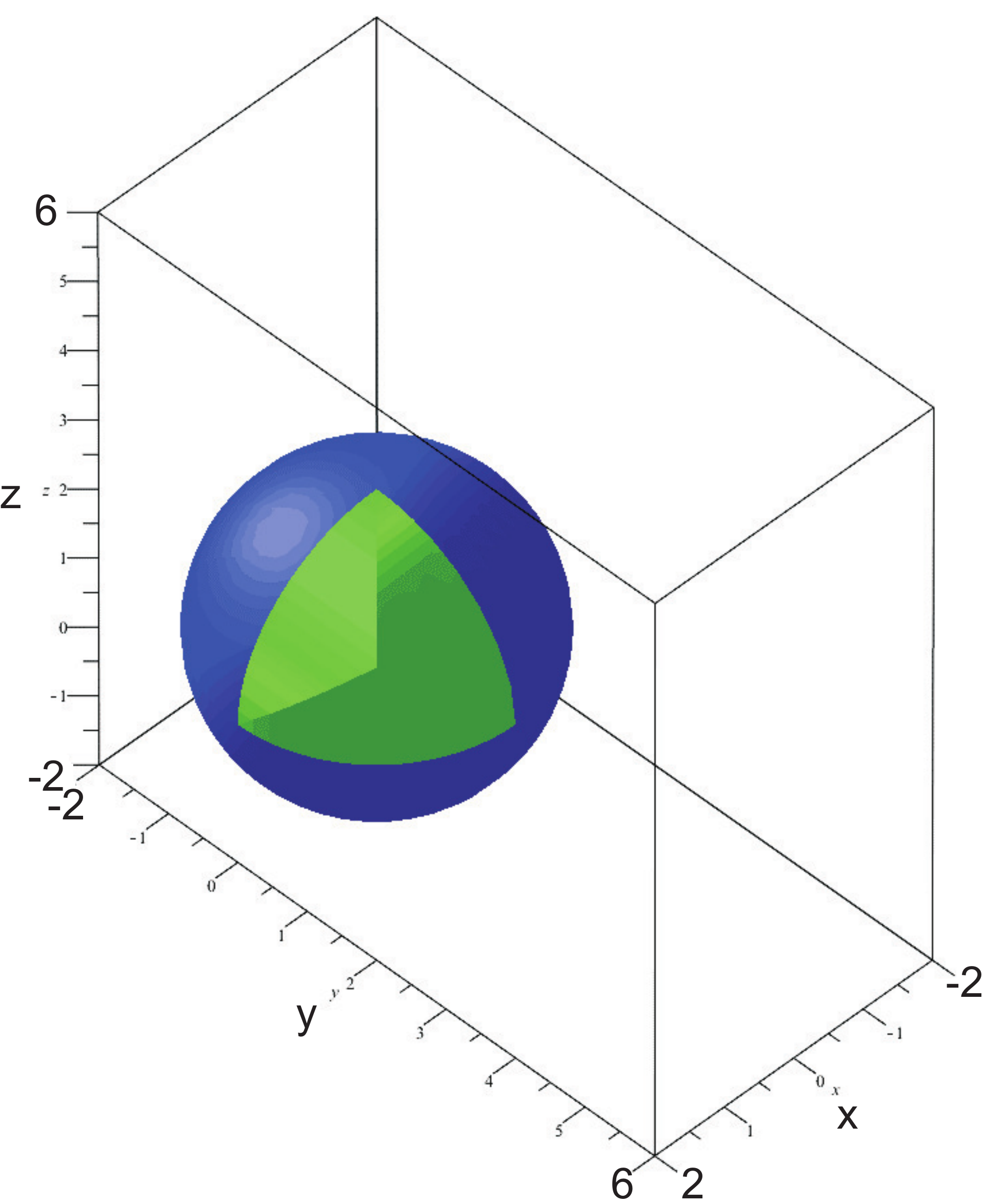}(a)
       \end{minipage} &
       \begin{minipage}[t]{45 mm}%
		    \psfrag{x}{$x$}
				\psfrag{y}{$y$}
				\psfrag{z}{$z$}
        \includegraphics[width=32 mm]{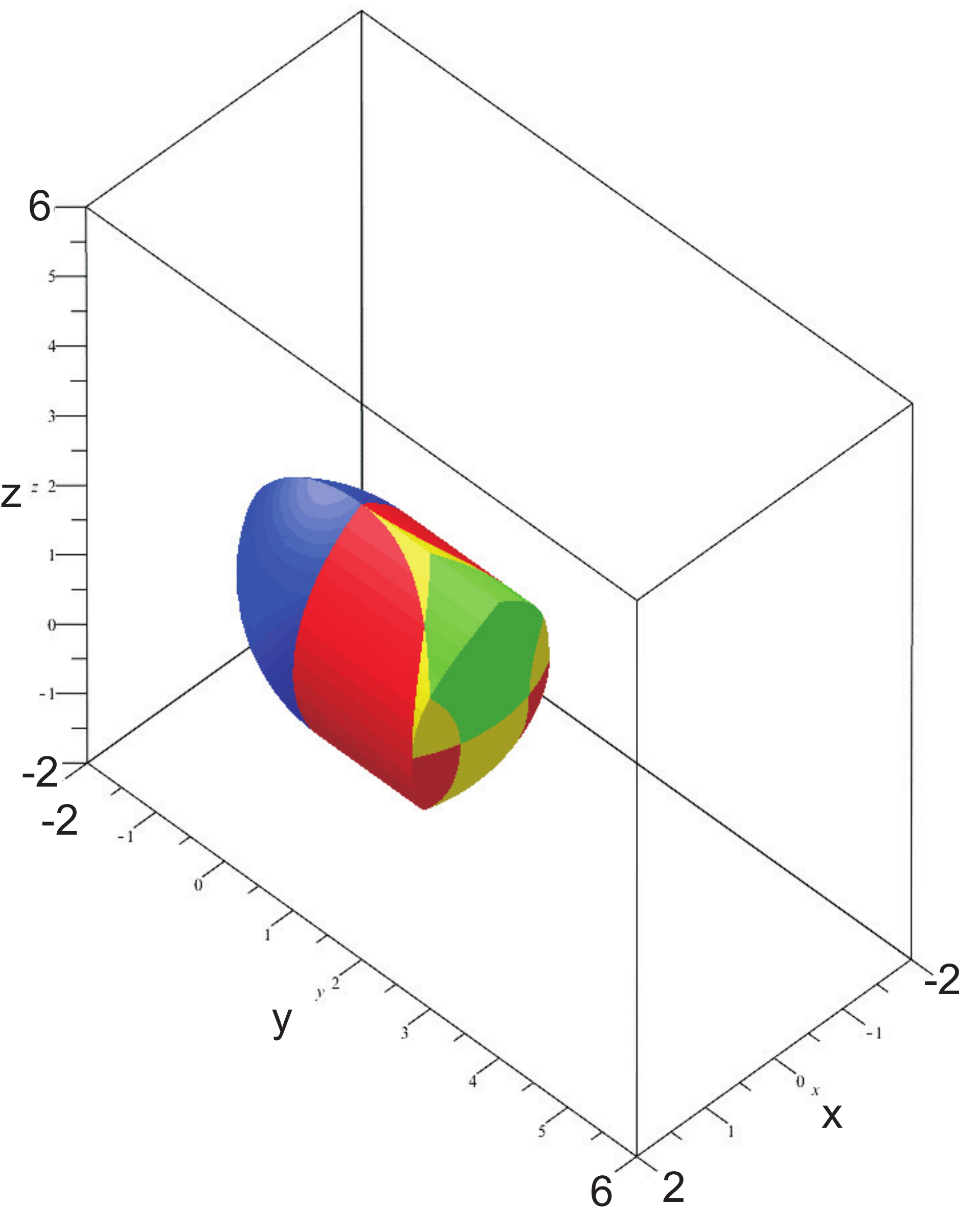}(b)
       \end{minipage} \\
       \begin{minipage}[t]{45 mm}%
		    \psfrag{x}{$x$}
				\psfrag{y}{$y$}
				\psfrag{z}{$z$}
        \includegraphics[width=32 mm]{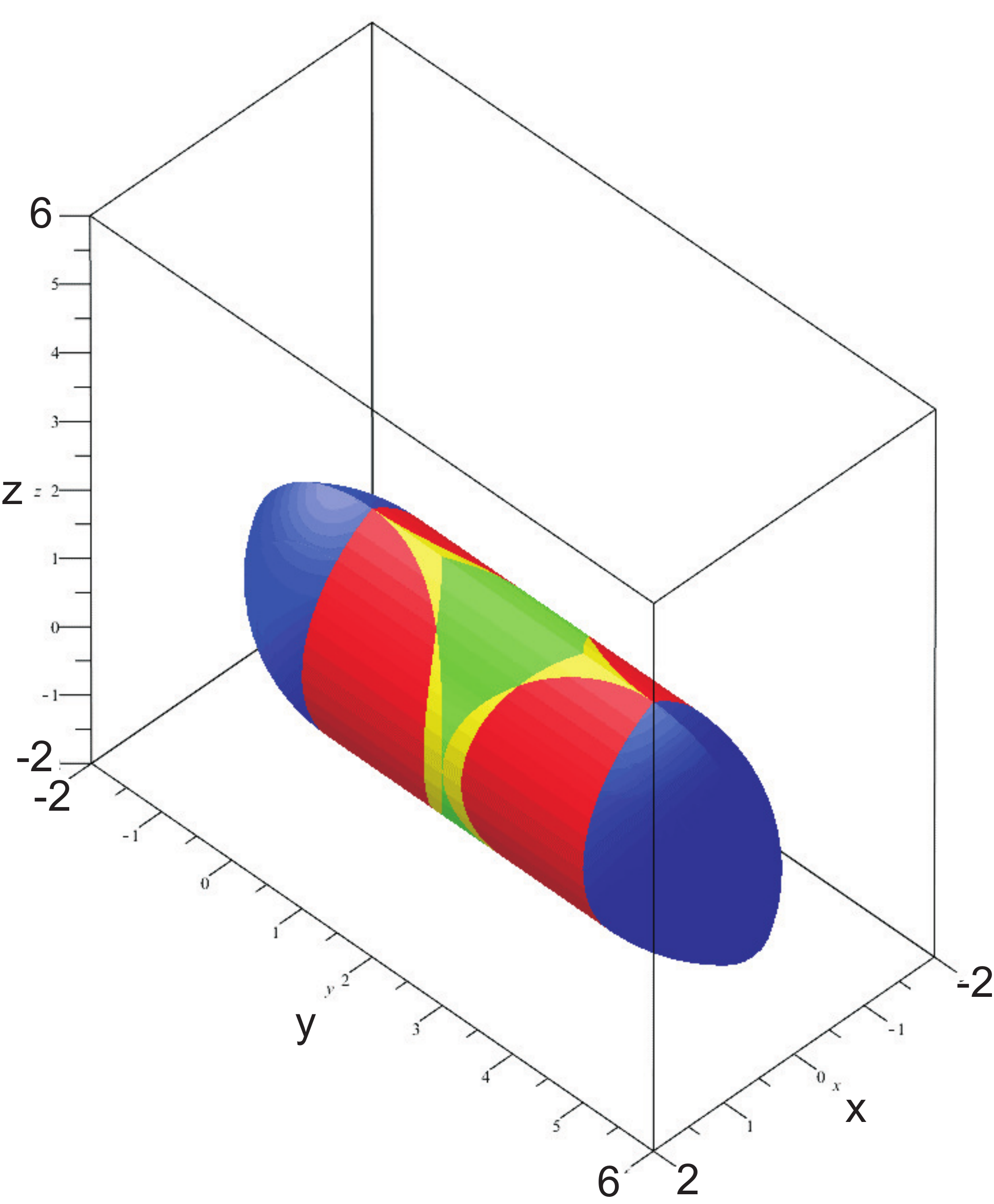}(c)
       \end{minipage} &
       \begin{minipage}[t]{45 mm}%
		    \psfrag{x}{$x$}
				\psfrag{y}{$y$}
				\psfrag{z}{$z$}
        \includegraphics[width=32 mm]{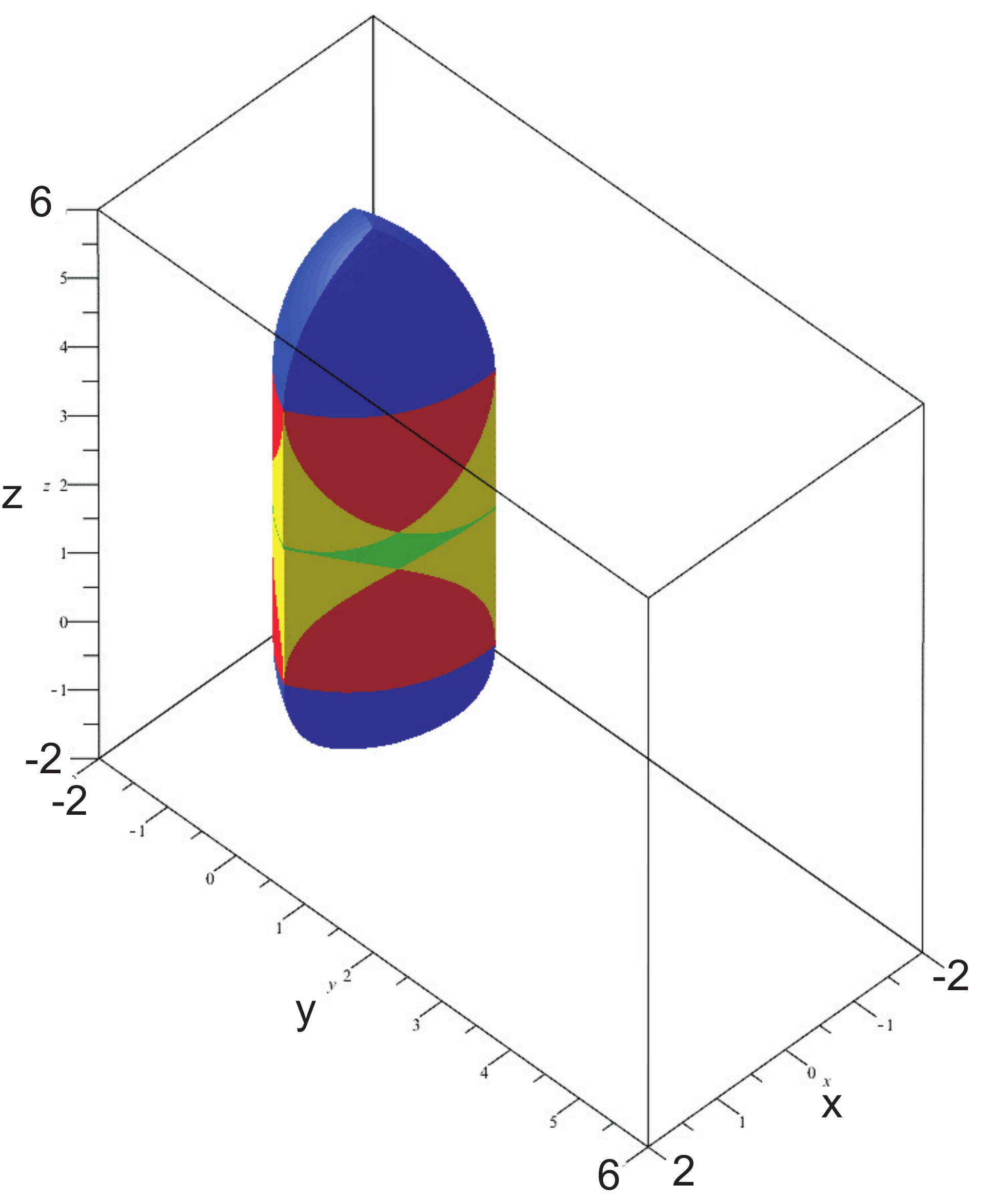}(d)
       \end{minipage}
    \end{tabular}
    \end{center}
    \caption[Workspace plot for delta robots]{Workspace with joint constraints for Orthoglide (a), Hybridglide (b), Triaglide (c) and UraneSX (d) robot with the number of inverse kinematic solutions. Blue, red, yellow and green regions corresponds to the one, two, four and eight  number of solutions for the IKP.}
    \protect\label{figure:workspace}
\end{figure}
\noindent The three main steps involved in the analysis are \cite{chab:2014, laz:2007, moroz:2008}:
\begin{itemize}
\item 	Computation of a subset of the joint space (resp. workspace) where the number of solutions changes: the {\it Discriminant Variety} .
\item 	Description of the complementary of the discriminant variety in connected cells: the Generic {\it Cylindrical Algebraic Decomposition} (CAD).
\item 	Connecting the cells belonging to the same connected component  in the counterpart of the discriminant variety: {\it interval comparisons.}
\end{itemize}
The different shapes of workspace for the delta-like robots is shown in Fig.~\ref{figure:workspace}, where blue, red, yellow and green regions correspond to the one, two, four and eight  number of
solutions for the IKP. A comparative study is done on the workspace of the family of delta-like manipulator and the results are shown in Fig.~\ref{figure:workspace}.  All the workspace are plotted in the rectangular box, where ${\bf x} \in [-2, 2 ]$, ${\bf y} \in [-2, 6 ]$ and ${\bf z} \in [-2, 6 ]$, so that the shapes of these workspace can be compared. From the Fig.~\ref{figure:workspace} it can be intuited that the Triaglide will be good selection, if the task space is more in horizontal plane, whereas the Orthoglide is good for the three dimensional task space.
\section{Conclusions}
A comparative study on the workspace of different delta-like robots gives the idea about shape of the workspace, which further plays an important role in the selection of the manipulator for the specific task or for the trajectory planning. The main characteristics associated with the singularities are tabulated in Table~\ref{complex}, which also gives some information about the complexity of the singularities, which is an essential factor for the singularity-free path plannings. From these data, it can be observed that the singularities associated with the Hybridglide are complicated, whereas the structure of those associated with the Triaglide is rather simple.
\begin{acknowledgement}
The work presented in this paper was partially funded by the Erasmus Mundus project ``India4EU II''.
\end{acknowledgement}

\end{document}